\documentclass{article}
\usepackage{arxiv}

\usepackage[utf8]{inputenc} 
\usepackage[T1]{fontenc}    
\usepackage{hyperref}       
\usepackage{url}            
\usepackage{booktabs}       
\usepackage{amsfonts}       
\usepackage{nicefrac}       
\usepackage{microtype}      
\usepackage{lipsum}
\usepackage{authblk}
\usepackage[misc]{ifsym}
\usepackage{graphicx}
\usepackage{multicol}
\usepackage{multirow}
\usepackage{amsmath}
\begin{document}
\title{Semi-supervised Breast Lesion Detection in Ultrasound Video Based on Temporal Coherence}
\author[1]{Sihong Chen}
\author[2]{Weiping Yu\thanks{This work was done when Weiping Yu and Xinlong Sun were interned at Tencent, Youtu X-lab Research. Sihong Chen and Weiping Yu were equal contribution.}}
\author[1]{Kai Ma}
\author[3]{Xinlong Sun}
\author[4]{Xiaona Lin}
\author[4]{Desheng Sun}
\author[1]{Yefeng Zheng\thanks{yefengzheng@tencent.com}}
\affil[1]{Tencent YouTu X-Lab, Shenzhen, China}
\affil[2]{School of Computer Science and Technology, Beijing Institute of Technology, Beijing, China}
\affil[3]{Tsinghua University, Shenzhen, China}
\affil[4]{Peking University Shenzhen Hospital, Shenzhen, China}
\renewcommand\Authands{ and }
\maketitle
\begin{abstract}
Breast lesion detection in ultrasound video is critical for computer-aided diagnosis. However, detecting lesion in video is quite challenging due to the blurred lesion boundary, high similarity to soft tissue and lack of video annotations. In this paper, we propose a semi-supervised breast lesion detection method based on temporal coherence which can detect the lesion more accurately. We aggregate features extracted from the historical key frames with adaptive key-frame scheduling strategy. Our proposed method accomplishes the unlabeled videos detection task by leveraging the supervision information from a different set of labeled images. In addition, a new WarpNet is designed to replace both the traditional spatial warping and feature aggregation operation, leading to a tremendous increase in speed. 
Experiments on 1,060 2D ultrasound sequences demonstrate that our proposed method achieves state-of-the-art video detection result as 91.3\% in mean average precision and 19 ms per frame on GPU, compared to a RetinaNet based detection method in 86.6\% and 32 ms.
\keywords{Semi-supervised \and Video detection \and Temporal Coherence \and Real-time \and Ultrasound \and Breast Lesion Detection.}
\end{abstract}

\section{Introduction}
\par
\begin{figure}
\begin{center}
\centerline{\includegraphics[width=0.95\textwidth]{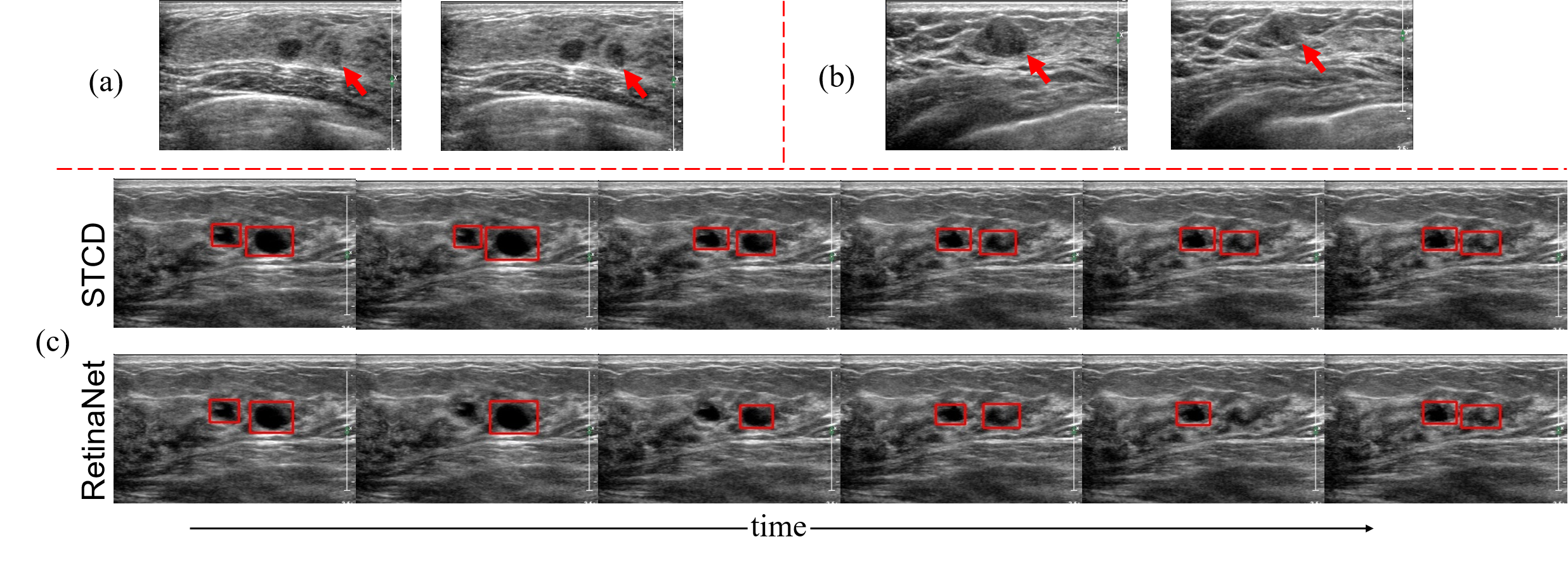}}
\end{center}
\caption{The challenging cases in breast lesion video detection. (a) Blurry boundaries; (b) High similarity to background; (c) Our proposed method on challenging samples}
\label{fig1}
\end{figure}

The 2D ultrasound is a widely used imaging modality for routine clinical diagnosis of breast lesions due to its advantages of real-time, low cost and non-invasion properties. However, ultrasound imaging highly depends on the sonographer's skill compared to other commonly used techniques such as mammography. Furthermore, interpreting ultrasound images requires an experienced and well-trained sonographer due to the complexity and presence of speckle noise and artifacts. Thus, Computer-Aided Diagnosis (CAD) could be beneficial to minimize the influence of the operator-dependent nature of ultrasound imaging and help sonographers in lesion detection. However, it's quite challenging to detect breast lesions in the individual frame of ultrasound video for the following reasons. Firstly, the lesion boundary is blurry in some frames which leads to confusion for the detector (see Fig.~\ref{fig1}a). Second, high similarity to the background of soft tissues and shadow artifacts result in failure detection of lesions (see Fig.~\ref{fig1}b). Moreover, there is only 2D image annotations because sonographers commonly save and label the key frames only. Without supervision information in video, training a video detection network becomes difficult.
\par
There are several existing methods for video detection. Methods based on correlation-filtering such as SiameseFC\cite{bertinetto2016fully} and kernelized correlation filters\cite{henriques2015high} process current frame with correlation filters which is related to the detection of the previous frame. If the detection of the previous frame is wrong, the features of non-target area propagated from the previous frame will affect the prediction result of the current frame. On the other hand, there is no video annotation in our dataset to support supervised SiameseFC method. Image-based detection methods such as YOLOV3\cite{redmon2018yolov3} and RetinaNet\cite{lin2017focal} can be trained with labeled 2D images and then be applied on the video data frame-by-frame. However, without considering temporal relationship of breast lesions, detection results on independent frame will be vulnerable to blurry boundary and interference from background which often result in wrong detection. Studies such as\cite{zhu2017deep,zhu2018towards} try to analyze temporal relationship by aggregating CNN features of the current frame and features extracted from FlowNet~\cite{dosovitskiy2015flownet,ilg2017flownet} that predicts the x-y flow fields. The original FlowNet can effectively analyze the temporal relationship but suffers from the time-consuming traditional warpping operation, and the gap between natural images and medical images since FlowNet leverages pre-trained weights from a natural image dataset.
\par
\begin{figure}
\begin{center}
\centerline{\includegraphics[width=0.9\textwidth]{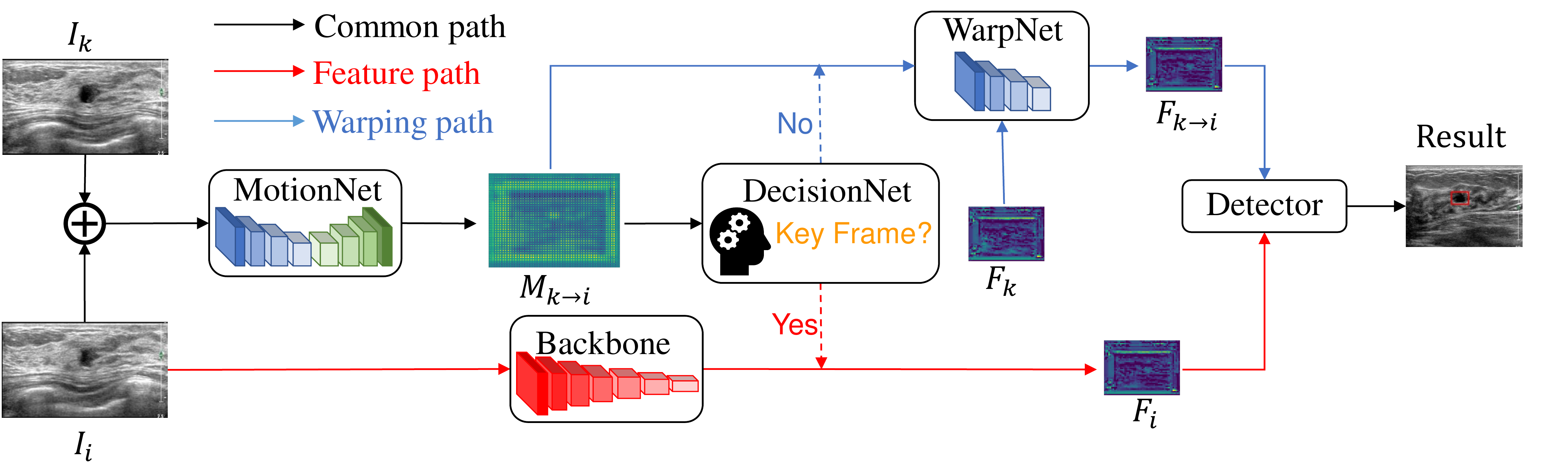}}
\end{center}
\caption{STCD Workflow} 
\label{fig2}
\end{figure}

In this work, we propose an end-to-end semi-supervised breast lesion detection network based on temporal coherence and we denote it as STCD (i.e., semi-supervised temporal coherent detection) for convenience.
In this work, we propose an end-to-end semi-supervised breast lesion detection network based on temporal coherence and we denote it as STCD for convenience. The STCD includes four parts as shown in Fig.~\ref{fig2}: the proposed MotionNet for calculating the difference between two frames, the DecisionNet~\cite{xu2018dvsnet} for judging whether the current frame is one of the key frames or not, the proposed WarpNet for transforming key-frame features to the current frame, and the backbone pre-trained RetinaNet~\cite{lin2017focal} for extracting CNN features and detecting.
\par
The proposed network makes three contributions as following: 1) STCD offers a semi-supervised solution to train a network with unlabeled video data and a different set of labeled images. 2) STCD improves breast lesion detection accuracy by automatically analyzing the temporal relationship across frames (see Fig.~\ref{fig1}c). 3) Due to the sparse distribution of key frames in the video, most of the video frames are processed by the proposed efficient MotionNet and WarpNet, which help the network achieve real-time performance.
\section{Methods}
\subsection{STCD Network}
\par
Since breast ultrasound video is commonly acquired with a certain structural direction at a relatively consistent pace, changes across frames are gradual and coherent most of the time. A frame which is different from adjacent frames in a specific time period is denoted as \emph{key frame} in this paper. Key frames in a sequence usually contain important information of the entire video. To be noted, the first frame in a video is always taken as the key frame.
\par
As shown in Fig.~\ref{fig2}, we extract features $F_{k}$ of the current key frame $I_{k}$ with the backbone network pre-trained from labeled images. Given such key frame and an input frame $I_{i}$, the MotionNet generates features $M_{k{\to}i}$ which encodes the motion difference between the two frames. Then the key-frame scheduling mechanism cooperated with the DecisionNet~\cite{xu2018dvsnet} is adopted to classify the input frame, as a key frame or non-key frame. DecisionNet predicts a score $S_{i}$ which indicates the consistency between $I_{k}$ and $I_{i}$. If $S_{i}$ is larger than a pre-defined threshold $\tau$, then $I_{i}$ is classified as the next key frame and processed by the backbone network to extract features. Otherwise, $I_{i}$ is treated as a non-key frame and the current key-frame features along with $M_{k{\to}i}$ are combined as the feature representations, as $I_{i}$ is similar to $I_{k}$.     
\par
In a video, for the sparse and specific key frames, a large backbone with excellent detection performance is used to extract features. For frames similar to the historical key frame, a lightweight MotionNet and WarpNet can generate features relying on key frame. STCD improves the ability of adaptive temporal coherent analysis and significantly increases the running speed.

\subsection{Semi-supervised Training Strategy}
\begin{figure}
\begin{center}
\centerline{\includegraphics[width=0.9\textwidth]{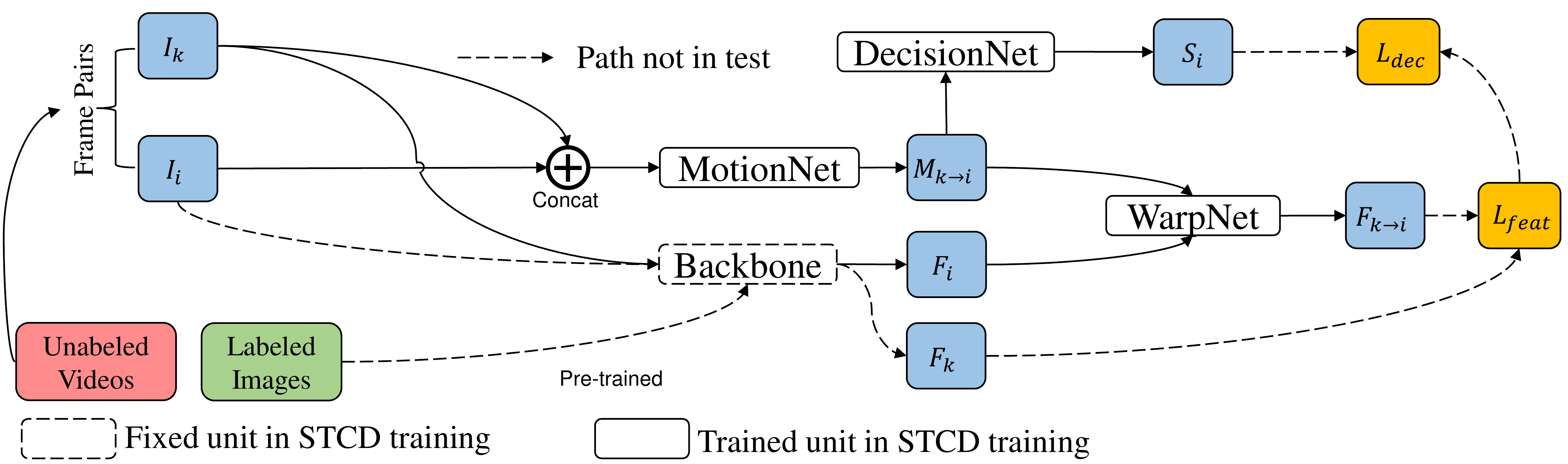}}
\end{center}
\caption{Semi-supervised Training of STCD} 
\label{fig3}
\end{figure}
\par
As shown in Fig.~\ref{fig3}, in order to utilize unlabeled videos and labeled images at the same time, we adopt a semi-supervised training strategy. To prepare training frame pairs composed of a key frame and a non-key frame for the STCD network, we randomly select a certain proportion of the frames as the key frames. Each of the above key frames is paired with a non-key frame in 20 adjacent frames. 
\par
During the training phase, we first train the detection network with the labeled image dataset. Then, we transfer the backbone of the detection network as the key-frame feature extractor without updating the weights. MotionNet, WarpNet, and DecisionNet are initialized with random parameters. At the beginning, features provided by MotionNet are not representative for breast lesions so that the DecisionNet based on MotionNet's features can not be effectively trained. Considering this phenomenon, after training MotionNet and WarpNet branches using a warm-up strategy for several epochs, we jointly train DecisionNet with other parts together to reach training convergence of STCD. 
\par
We also design two loss functions: feature loss $L_{feat}$ to measure correlation between $F_{k{\to}i}$ and $F_{i}$, and decision loss $L_{dec}$ to achieve adaptive key-frame scheduling mechanism. The correlation score $Q_{k{\to}i}$ which calculates correlation between $F_{k{\to}i}$ and $F_{i}$ can be defined as:
\begin{equation}
Q_{k{\to}i}=\frac{\sum(F_{k{\to}i}(p)-F_{i}(p))^{2}}{N_{p}}
\end{equation}
where $N_{p}$ is the total number of pixels in the feature maps, and $p$ is the index of a pixel.
$L_{feat}$ calculates mean values of series of $Q_{k{\to}i}$ in a batch, which forces features extracted from MotionNet and WarpNet to be close to features extracted from the labeled images. 
$L_{dec}$ enforces the network's output $S_{i}$ to be close to $Q_{k{\to}i}$ which can be summarized as:
\begin{equation}
L_{dec}=(Q_{k{\to}i}-S_{i})^2.
\end{equation}

\subsection{MotionNet and WarpNet}
As an important addition to STCD, MotionNet and WarpNet are mainly used to extract features of non-key frames. In a video, most of the frames are non-key frames whose detection accuracy and speed affect the detection results of the entire video significantly. Inspired by ~\cite{dosovitskiy2015flownet}, we design the MotionNet to calculate the motion between two frames. Considering the limited training data and inference efficiency, we reduce the output channel of each layer in the original FlowNetS~\cite{dosovitskiy2015flownet} structure to 1/4. MotionNet with less parameters effectively prevents over-fitting and increases speed. On the contrary,
traditional feature transformation methods~\cite{zhu2018towards,zhu2017deep} transform all the pixels of the entire feature map pixel-by-pixel with the displacement information. Such process is time-consuming due to the large number of feature map pixels. 
\par
Regarding the above shortcomings, we design a lightweight WarpNet to complete the transformation of the feature map.
First, the input feature map is adaptively resized to the same size of the target feature map via a 3$\times$3 convolutional operation. Subsequently, a weighted combination operation is performed by a 1$\times$1 convolution operation. The lightweight WarpNet transforms feature map in a learnable non-linear style which improves both the accuracy and efficiency. 
\section{Experiments}
\subsection{Datasets and Training}
We collected 5,608 labeled breast lesion ultrasound images, 80 unlabeled sequences and 10 labeled sequences from 90 patients, where each video sequence contains 105 to 188 frames. Labeled ultrasound images are used to train the detection network such as RetinaNet~\cite{lin2017focal} and YOLOV3~\cite{redmon2018yolov3}. Together with 80 unlabeled videos, a hybrid dataset is used to train STCD and 10 labeled sequences are used to be the test set. We select $1/3$ frames of the unlabeled training sequences as the key frames to form 59,962 training pairs. We train STCD with Adam~\cite{kingma2014adam} optimizer. In the first 10 epochs, we set the learning rates of DecisionNet, MotionNet, and WarpNet as 0, 0.001, and 0.001, respectively and weights are reduced by 10\% after each epoch. After 10 epochs, we change weight of DecisionNet to 0.001 to turn it on. We conduct experiments on both CPU (Intel Xeon E5-2680) and GPU (Nvidia Tesla P40).
\subsection{Effectiveness of STCD}
\begin{figure}
\begin{center}
\centerline{\includegraphics[width=0.8\textwidth]{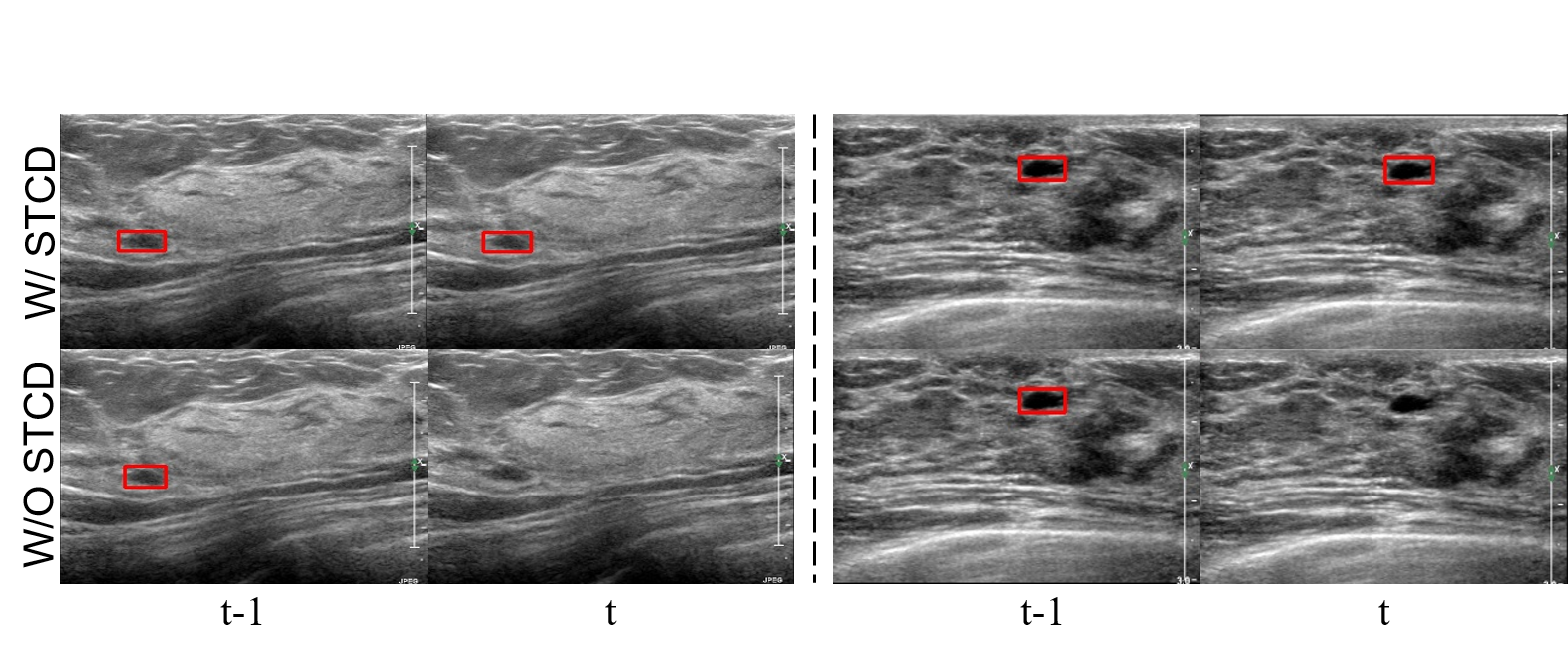}}
\end{center}
\caption{Breast lesion detection results of RetinaNet and the proposed STCD method} 
\label{fig7}
\end{figure}

In this section, we design several experiments to quantitatively (see Table~\ref{tab1}) and qualitatively (see Fig.~\ref{fig7}) evaluate the effectiveness and efficiency of different backbone networks. 
As illustrated in Table~\ref{tab1}, STCD with different backbones improves the accuracy in a range between 1.5\% and 5.4\%, GPU speed increment between 32\% and 68\% and CPU between 69\% and 151\%. Together with the qualitative results shown in Fig.~\ref{fig7}, we can see that STCD brings more historical target structural information by temporal coherence analysis compared to the frame-based detection. 
On the other hand, as the depth of network increases, the accuracy and efficiency improvement from STCD increases. This shows that the features extracted from the key frames by the large network are more effective which can further improve detection performance of non-key frames. Therefore, we select RetinaNet with ResNet-152 as detection backbone in our proposed STCD.
\begin{table}
\begin{center}
\caption{Performance of STCT with different backbones for breast lesion detection on the test set of ultrasound videos}
\centering
\begin{tabular}{c|c|c|c|c}
\hline
Method & Backbone & mAP(\%) & GPU Runtime (ms) & CPU Runtime (ms)\\
\hline
\hline
YOLOV3~\cite{redmon2018yolov3} & DarkNet & {80.6} & 25 & 1071 \\ \hline
YOLOV3 (STCD) & DarkNet & {\bfseries81.8} & {\bfseries{19}} & {\bfseries{630}} \\ \hline\hline
RetinaNet~\cite{lin2017focal} & ResNet-50 & {82.0} & 23 & 913 \\ \hline
RetinaNet (STCD) & ResNet-50 & {86.0} & {\bfseries{16}} & {\bfseries{540}} \\ \hline
RetinaNet~\cite{lin2017focal} & ResNet-101 & {85.4} & 28 & 1161 \\ \hline
RetinaNet (STCD) & ResNet-101 & {89.9} & 17 & 577 \\ \hline
RetinaNet~\cite{lin2017focal} & ResNet-152 & {86.6} & 32 & 1548 \\ \hline
RetinaNet (STCD) & ResNet-152 & {{\bfseries91.3}} & 19 & 619 \\ \hline
\end{tabular}
\label{tab1}
\end{center}
\end{table}
\par
\begin{figure}
\begin{center}
\centerline{\includegraphics[width=0.3\textwidth]{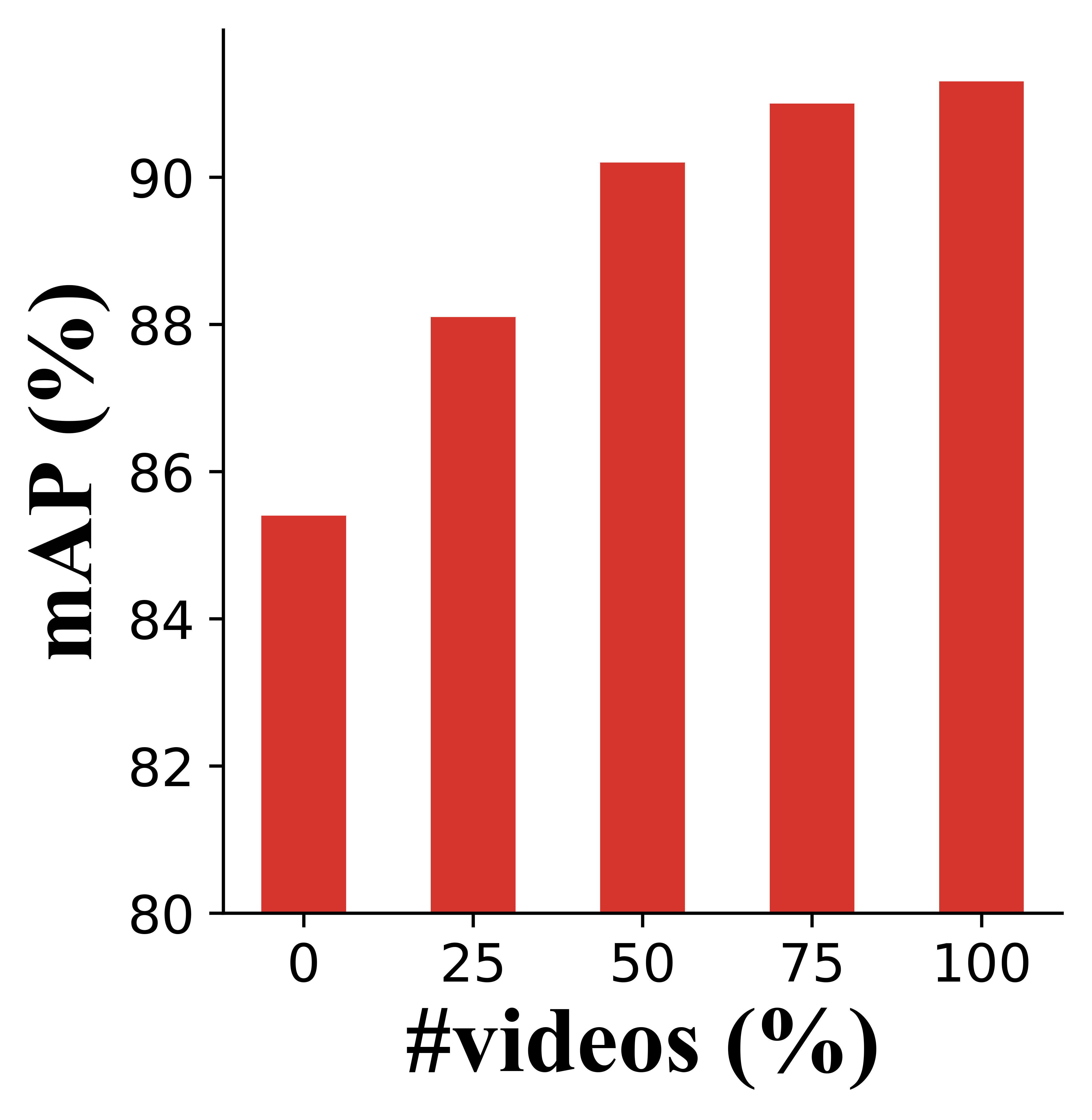}}
\end{center}
\caption{Effectiveness of Unlabeled Data} 
\label{fig4}
\end{figure}

To further evaluate the impact of unsupervised data volume on performance, we used different amounts of unlabeled data to train STCD for comparison in Fig.~\ref{fig4}. As shown in Fig.~\ref{fig4}, as the amount of unsupervised data increases, the detection accuracy gradually increases. Such phenomenon demonstrates that the unlabeled video data brings useful information for video detection using the semi-supervised mechanism introduced by STCD.

\subsection{Effectiveness of WarpNet and MotionNet}
\begin{figure}
\begin{center}
\centerline{\includegraphics[width=0.77\textwidth]{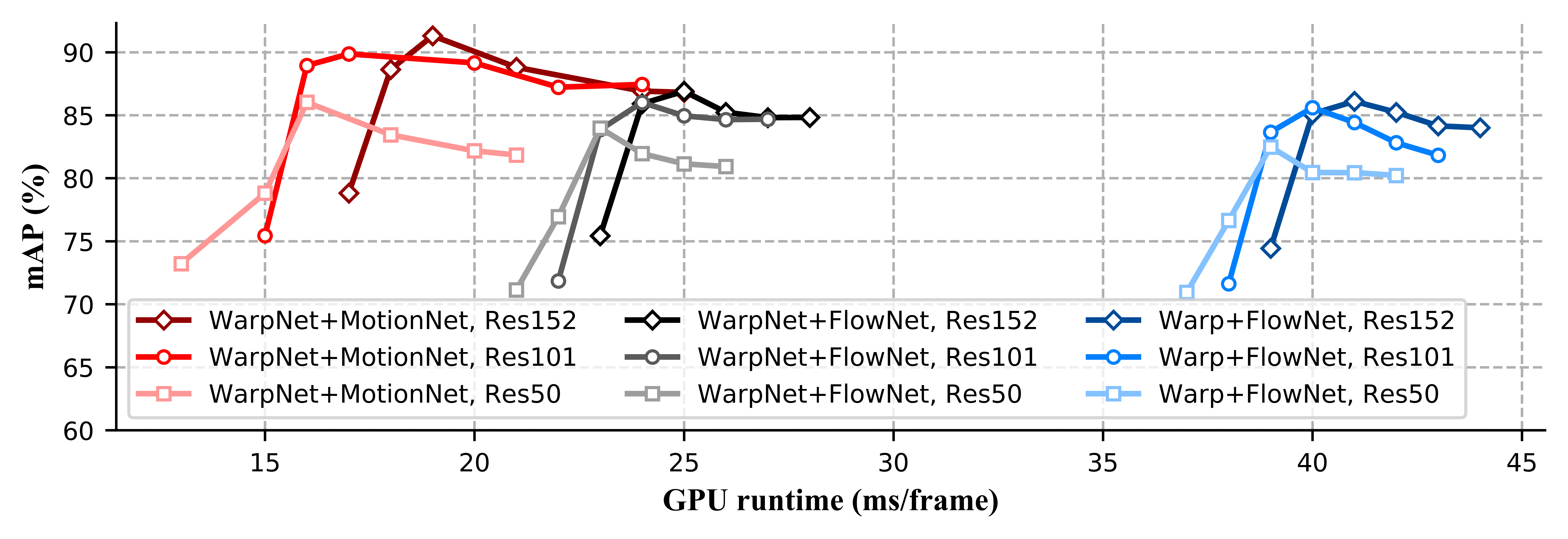}}
\end{center}
\caption{Accuracy (mAP) and runtime of breast lesion detection under different threshold $\tau$ for key frame decision} 
\label{fig5}
\end{figure}

We design three sets of comparative experiments: proposed MotionNet with WarpNet, MotionNet with traditional warp transformation, and FlowNet~\cite{dosovitskiy2015flownet} with transformation. As shown in Fig.~\ref{fig5}, with different decision thresholds for key frames, both MotionNet and WarpNet can bring accuracy improvement compared to FlowNet and Warp operation. 
It shows that MotionNet and WarpNet that are based on an end-to-end network are more adaptive. Furthermore, FlowNet highly relies on pre-trained weights from Flying Chairs natural image dataset~\cite{dosovitskiy2015flownet} which is different from medical images. Compared to FlowNet and untrainable warp operation, MotionNet and WarpNet can obtain more breast lesion information during semi-supervised training so as to improve detection accuracy. Since WarpNet can adaptively transform features with limited CNN parameters, WarpNet brings nearly a 1x speed increase over the previous Warp operation. The compressed MotionNet also has a speed increase of approximately 40\% compared to FlowNet.
\subsection{Ablation Study}
To further explore the impact of different key-frame scheduling and architectures of MotionNet, two sets of experiments are conducted as follows: 1) we compare our proposed key-frame scheduling method based on DecisionNet with the fixed key frame selection stragety~\cite{zhu2017deep} (Fixed), key frame selection stragety based on mean squared error between two grey images~\cite{xu2018dvsnet} (Grey Correlation) and strategy based on optical-flow feature map between two images (Flow-guided Correlation); and, 2) we compare our MotionNet concatenating with MotionNet-b based on FlowNetC~\cite{dosovitskiy2015flownet} and MotionNet-c based on FlowNet2~\cite{ilg2017flownet}. Both MotionNet-b and MotionNet-c are compressed by cutting half number of channels in each layer. For training efficiency, both experiments are performed on RetinaNet with ResNet-101.
\par
\begin{figure}
\begin{center}
\centerline{\includegraphics[width=0.77\textwidth]{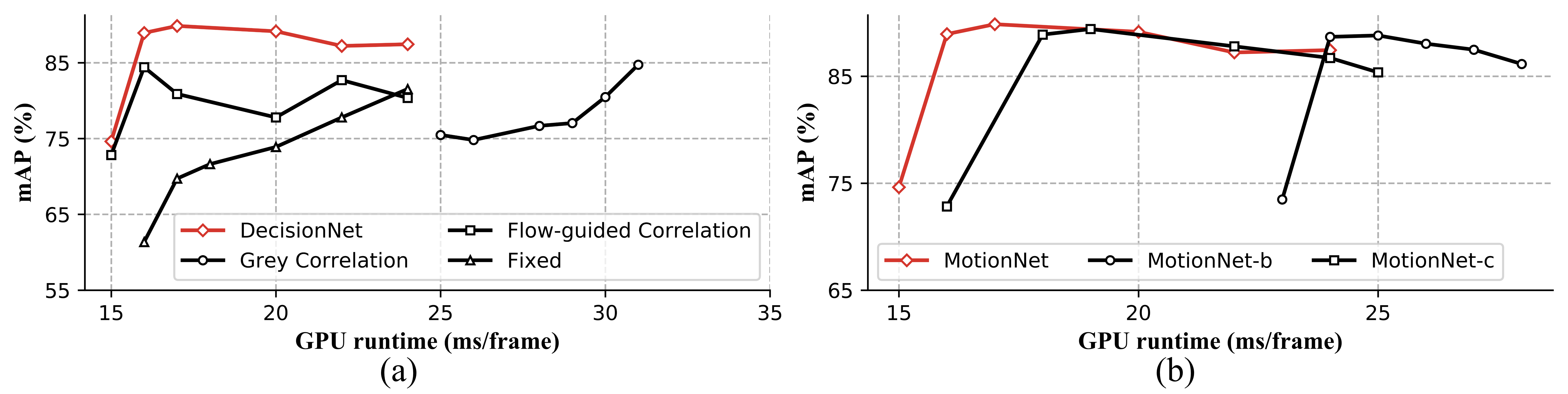}}
\end{center}
\caption{Accuracy (mAP) and runtime of (a) different key-frame scheduling; (b) different MotionNet architectures under different threshold $\tau$ for key frame decision} 
\label{fig6}
\end{figure}

As shown in Fig.~\ref{fig6}a, our approach surpasses other methods both in accuracy and speed. Furthermore, our method has a 23\% increase in accuracy compared to the Fixed strategy, and a 61\% increase in speed compared to the Grey Correlation approach, which demonstrates that our method can adaptively find target difference between two frames, resulting in better yet less key frames so as to improve video detection accuracy and efficiency. As our proposed method introduces the detection supervision information extracted by pre-trained detection backbone, it leverages more spatially specific information which can improve detection accuracy.
In addition, as shown in Fig.~\ref{fig6}b, the proposed MotionNet has slightly improved accuracy and speed compared to the MotionNet-b with reduced convolutional parameters. Compared with the MotionNet-c, it has about 1.3\% increase in accuracy and a 50\% increase in speed. On the contrary, MotionNet-c contains a large number of repetitive modules, and the amount of network parameters has increased dramatically. With limited training data, it is easy to cause over-fitting and performance degradation.

\section{Conclusion}
In this work, we proposed the STCD network to detect breast lesions in ultrasound videos. By employing the coherence between adjacent frames, STCD can improve the detection accuracy with assistant of information from the historical frame. The semi-supervised learning approach introduced more video information and improved detection accuracy. Meanwhile, runtime efficiency was also improved significantly.

\bibliographystyle{unsrt}

\end{document}